\documentclass{article} 
\usepackage{colm2024_conference}

\usepackage{microtype}
\usepackage{hyperref}
\usepackage{url}
\usepackage{booktabs}
\definecolor{darkblue}{rgb}{0, 0, 0.5}
\hypersetup{colorlinks=true, citecolor=darkblue, linkcolor=darkblue, urlcolor=darkblue}
\usepackage{graphicx} 
\usepackage{tcolorbox}
\usepackage{colortbl,booktabs}
\usepackage{enumitem}
\usepackage{tikz}
\usepackage{multirow}
\usepackage{color}
\usepackage{xcolor}
\usepackage{colortbl,booktabs}
\usepackage{graphicx,wrapfig,lipsum}
\usepackage{fontawesome}
\usepackage[ruled,vlined]{algorithm2e}

\SetCommentSty{mycommfont}
\usepackage{wrapfig}
\SetKwInput{KwInput}{Input}                
\SetKwInput{KwOutput}{Output}              

\newcommand*\circled[1]{\tikz[baseline=(char.base)]{
            \node[shape=circle,draw,inner sep=0.4pt] (char) {#1};}}

\title{FFN-SkipLLM: A Hidden Gem for Autoregressive Decoding \\ with Adaptive Feed Forward Skipping}

\author{Ajay Jaiswal$^1$, Bodun Hu$^1$, Lu Yin$^{2,5}$,  Yeonju Ro$^1$,  \textbf{Shiwei Liu$^{2,4}$,  Tianlong Chen$^3$,  Aditya Akella$^1$} \\$^1$University of Texas at Austin\\ $^2$Eindhoven University of Technology \\$^3$University of North Carolina at Chapel Hill \\$^4$University of Oxford \\$^5$University of Aberdeen}

\colmfinalcopy 
\begin{document}

\maketitle

\begin{abstract}
Autoregressive Large Language Models (\emph{e.g.,} LLaMa, GPTs) are omnipresent achieving remarkable success in language understanding and generation. However, such impressive capability typically comes with a substantial model size, which presents significant challenges for autoregressive token-by-token generation. To mitigate computation overload incurred during generation, several \textit{early-exit} and \textit{layer-dropping} strategies have been proposed. Despite some promising success due to the redundancy across LLMs layers on metrics like Rough-L/BLUE, our careful knowledge-intensive evaluation unveils issues such as generation collapse, hallucination of wrong facts, and noticeable performance drop even at the trivial exit ratio of $\sim$ 10-15\% of layers. We attribute these errors primarily to ineffective handling of the KV cache through state copying during early-exit. In this work, we observed the \textit{saturation of computationally expensive feed-forward blocks} of LLM layers and proposed \textbf{FFN-SkipLLM}, which is a novel fine-grained skip strategy of autoregressive LLMs. More specifically, FFN-SkipLLM is an \textit{input-adaptive feed-forward skipping strategy} that can skip $\sim$ 25-30\% of FFN blocks of LLMs with marginal change in performance on knowledge-intensive generation tasks without any requirement to handle KV cache. Our extensive experiments and ablation across benchmarks like MT-Bench, Factoid-QA, and variable-length text summarization illustrate how our simple and ease-at-use method can facilitate a faster autoregressive decoding. 
\end{abstract}

\section{Introduction}
Autoregressive Large Language Models (LLMs) have been recently \textit{show-stealer}, profoundly influencing not only the landscape of NLP \citep{ram2023context,liu2023llmrec,sawada2023arb,jaiswal2021radbert,qin2023chatgpt,zhuo2023large,Lee2023CanLL}, but also recently buttressing numerous computer vision \citep{lian2023llm,wang2023visionllm,lai2023lisa,lu2023chameleon,li2024cancergpt} and graph neural networks \citep{ye2023natural,chen2023exploring,qian2023can,duan2023simteg,chen2024llaga} algorithms; achieving steller performance across various task benchmarks. However, their widespread adoption is hindered by their massive scale, characterized by billions of parameters, which demand exceedingly high computational resources and memory capacities. For instance, the GPT-175B model necessitates 325 GB of GPU memory for loading its weights and relies on a minimum of five A100 (80GB) GPUs employing sophisticated parallelism techniques \citep{sheng2023high}. This imposing computational and memory requirement presents a challenge to the broader accessibility of these models.

To alleviate the demanding hardware requirements for deploying massive trained models, considerable efforts have been taking to mitigate their high computational inference cost resulting from token-by-token generation. Among several model compression techniques such as quantization \citep{liu2023llm,Kim2023MemoryEfficientFO,Dettmers2023QLoRAEF,Frantar2022GPTQAP,Lin2023AWQAW,Dettmers2023SpQRAS}, and sparse neural networks \citep{frankle2018the,chen2020lottery,jaiswal2022training,lee2018snip,zhangheng2023sparse,jaiswal2023instant,jaiswal2023emergence,liu2023sparsity,Yin2023PruningSP,yin2023outlier} which require additional hardware support for speedup, \textit{token-level early exit or layer-skip} has emerged as a promising technique to alleviate these limitations by allowing tokens to cease computation as soon as their hidden states reach saturation \citep{hash-skip-sun,del2023skipdecode,calm,men2024shortgpt}. These methods exploit existing redundancy across LLMs layers which can be ignored during token-by-token generation significantly saving massive computation involved within a layer (\emph{e.g.,} $\sim$ 200-300 million parameters in a single LLaMa layer). Although the proposed methods have shown some promising success, their performance is widely restricted by the issue of inappropriately handling KV caching. KV caching saves keys and values of all attention layers for previously generated tokens and accelerates sequence generation by reducing redundant computation (though at the cost of higher memory usage). Given a token is generated via early exiting, its KV caches in subsequent layers are incomplete which impedes the generation of future tokens beyond the exiting layer of the current token.

\begin{wraptable}{r}{5cm}
\vspace{-1em}
\scriptsize
    \centering
    \begin{tabular}{ccc}
    \toprule
        Layer Name & \# Parameters \\
    \midrule
        attention.wq.weight & $\sim$ 16.77M\\
        attention.wk.weight  & $\sim$ 16.77M\\
        attention.wv.weight  & $\sim$ 16.77M\\
        attention.wo.weight  & $\sim$ 16.77M\\
        feed\_forward.w1.weight  & $\sim$ 45.08M\\
        feed\_forward.w2.weight & $\sim$ 45.08M\\
        feed\_forward.w3.weight & $\sim$ 45.08M \\
    \bottomrule
    \end{tabular}
    \vspace{-1em}
    \caption{Parameter counts of LLaMa-7B layer component.}
    \label{tab:parameter_count}
    \vspace{-1em}
\end{wraptable}

\begin{figure}
    \centering
    \includegraphics[width=0.99\linewidth, trim= 1em 1em 1em 1em]{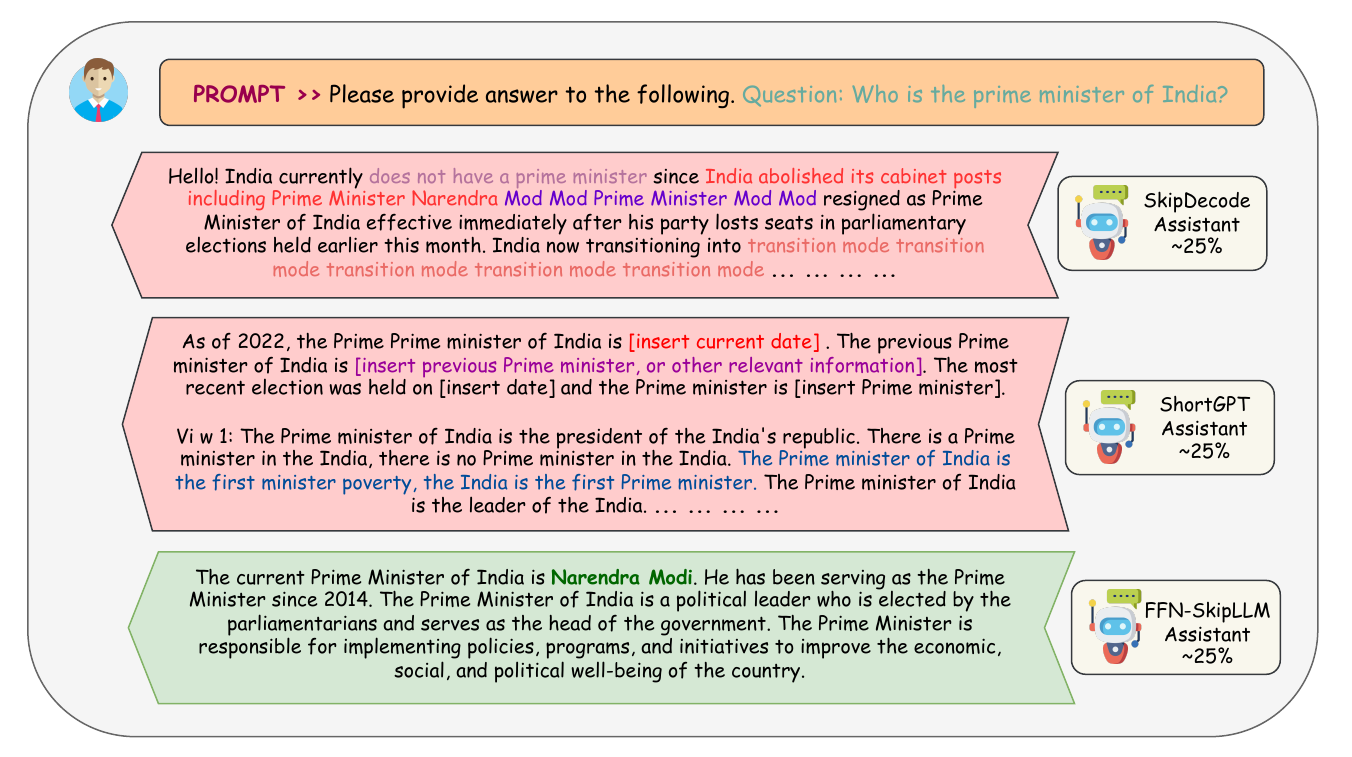}
    \caption{\textbf{Merits of Autoregressive Decoding with Layer Skipping:} Comparison of the responses generated by two recent Layer Skipping methods, namely SkipDecode \citep{del2023skipdecode} and ShortGPT \citep{men2024shortgpt} for a knowledge-intensive QA example. It can be observed that both LLaMa-chat-13B model with $\sim$ 25\% layers skipped per token using SkipDecode and ShortGPT suffers from \textit{hallucination} and \textit{token collapse} (repetitive generation) while FFN-SkipLLM can still retrieve the correct response.}
    \label{fig:ffn_skipllm_motivation}
    \vspace{-1.5em}
\end{figure}

For handling KV cache issue, some recent works \citep{elbayad2019depth,calm,li2021accelerating,chen2023ee,del2023skipdecode} proposes three main solutions: copying hidden states, pre-fixed token-level skip pattern, and KV recomputation. Despite these mitigation methods, our careful knowledge-intensive investigation reveals that layer-skipping induces permanent damage due to deviation from the inference process that the model is trained to excel at, leading to significant hallucination of wrong facts and token generation collapse. Figure \ref{fig:ffn_skipllm_motivation} shows the comparison of the responses generated by two recent Layer Skipping methods, namely SkipDecode \citep{del2023skipdecode} and ShortGPT \citep{men2024shortgpt} for a knowledge-intensive QA example. In response, both ShortGPT and SkipDeocde fail to generate the correct answer ``Narendra Modi" and suffer from token collapse and hallucinate misinformation. 



In this work, we ask an interesting unexplored question:  \textit{Instead of attempting to fix the KV cache, can we completely circumvent the KV cache bottleneck of layer-skipping and still ignore unnecessary computational expenses while mitigating hallucination and token generation collapse?} To this end, our work is the \underline{\textbf{first}} attempt to investigate a fine-grained layer-skipping strategy that focuses on computationally expensive feed-forward network (FFN) blocks in LLMs. Table \ref{tab:parameter_count} presents the parameter counts of individual components of LLaMa-7B layer and it can be observed that FFN blocks hold approximately \textit{two-third} of the parameter budget of the layer, marking them as favorable candidates for skipping during token-by-token generation. Our work derives its motivation from two primary observations: \circled{1} we find a \textit{monotonically increasing cosine similarity} between the tensors generated before and after the FFN blocks across layers in LLMs which indicates unnecessary computation performed by these blocks, \circled{2} due to the observed phenomenon of attention sink \citep{xiao2023efficient}, we found that allowing a \textit{small fraction of first-few token ($\sim$ 5-10\% of maximum sequence length) decoding using full strength} (no-skip) of LLMs can significantly help in stabilizing the KV cache, paving way for skipping FFN blocks without significant performance degradation for later tokens. We propose \textbf{FFN-SkipLLM},  a novel fine-grained skip strategy of autoregressive LLMs which is an input-adaptive
feed-forward skipping strategy that can skip $\sim$ 25-30\% of FFN blocks of LLMs
with marginal change in performance on knowledge-intensive tasks. Note that because we only skip FFN blocks, we in-turn can fully circumvent the KV cache issue associated with layer-skipping. Our primary contributions can be unfolded as:

\begin{itemize}
    \item Unlike prior layer-skipping methods, we focus on only skipping computationally expensive FFN blocks based on our observation of their monotonically increasing saturation within the middle layers of LLMs.

    \item Our proposed FFN-SkipLLM uses a simple \textit{cosine similarity} metric across tensors to capture the trend of FFN saturation and decide an input-adaptive skipping of FFN blocks. More specifically, once a similarity threshold is reached, given the monotonically increasing saturation, we \textbf{greedily} select the next $k$ layers whose FFN blocks can be ignored depending on the desired skipping requirement.   

    \item Our extensive knowledge-intensive experiments such as Factoid-QA, Multi-turn conversations and Variable-length in-context text summarization, reveal that FFN-SkipLLM can skip $\sim$ 25-30\% of FFN blocks of LLMs with a marginal change in performance and reduce hallucination and token collapse. 
\end{itemize}




\section{Layer-skipping: An Knowledge-Intensive Evaluation}
Recent advancements in autoregressive models  \citep{touvron2023llama, qin2023chatgpt, zhang2022opt} have revolutionized the quality of language generation in various generative tasks, including question answering \citep{rajpurkar2016squad}, summarization \citep{fabbri2019multi, nallapati2016abstractive}, and machine translation \citep{bahdanau2014neural}. However, these large transformer models face challenges in terms of high inference latency attributed to their numerous layers and the autoregressive decoding process. The sequential computation of multiple stacks of transformer layers for each token during the inference stage imposes significant computational overheads, thus limiting their real-time adaptability.

To counter the computational cost of token-by-token generation with modern gigantic LLMs, several works \citep{Chen2023EELLMLT,men2024shortgpt,del2023skipdecode,kim2024shortened,bae2023fast} have been recently exploring token-level early exit and layer-skipping (depth-pruning) strategies. The primary challenge associated with these approaches is that if the current token exits at a higher layer, there arises a need to recalculate the Key-Value (KV) caches for preceding tokens. This mandatory recalibration increases the computational burden and diminishes the benefits of early exit techniques, as the computation of each preceding token becomes contingent on the computation of subsequent tokens. To this end, three major approaches has been explored: (1) copy the hidden states of the current token at the exiting layer to all later layers, which will be used to compute the keys and values at later attention layers, (2) pre-specify the exiting layer for each token, while ensuring that KV missing in previous tokens will not hinder the generation of later tokens; with this approach, the ability of token-wise adaptive selection of exits is inevitably lost, (3) KV recomputation which is a variant of synchronized parallel decoding and adds additional computational and memory overhead. 

Despite some notable performance gains over some metrics (\emph{e.g.}, perplexity, Rough-L, BLUE), our careful knowledge-intensive investigation reveals that the KV cache problem during layer-skip is not effectively addressed. Figure \ref{fig:ffn_skipllm_motivation} illustrates the responses generated by two recent layer-skipping methods SkipDeocde \citep{del2023skipdecode} and \citep{men2024shortgpt} for a given factoid-based QA task which requires answering using relevant entities and attributes ingested within LLMs during pre-training. Interestingly, answers generated by the SkipDecode agent hallucinate misinformation claiming \textit{`... does not have a prime minister ... India abolished its cabinet posts ... `} while the ShortGPT agent fails to generate any factoid to answer the question. Note that both agents suffer from token collapse and start generating repetitive content after some time. To quantitatively estimate the damage of layer-skipping, Table \ref{tab:knowledge_evaluation} presents the performance of SkipDecode and ShortGPT with respect to the full model on three knowledge-rich tasks (Section \ref{sec:factoid-qa}, \ref{sec:summarization}, \ref{sec:mt-bench}) that closely resemble the daily use-cases for LLMs. It can be observed that despite impressive results reported on traditional metrics, we found the performance significantly suffers when compared to the full model. To this end, in this work, we attempt to explore an orthogonal direction that diverge from conventional layer-skipping and investigate the potential of skipping computationally heavy FFN blocks across layers which accounts for approximately two-third of the parameter count.

\begin{table}[]
\scriptsize
    \centering
    \begin{tabular}{c|ccc}
    \toprule
        \textbf{Method ($\sim$ 20\% Skip)} & \textbf{Factoid-QA} & \textbf{Multi-turn Conversation} & \textbf{In-context Summarization} \\
    \midrule
        Full Model & 79.02 & 7.61 & 8.15\\
        SkipDecode \citep{del2023skipdecode} & 73.33 & 6.53 & 7.47\\
        ShortGPT   \citep{men2024shortgpt} & 70.49 & 6.17 & 6.33\\
    \midrule
        \rowcolor[gray]{0.9}
        Ours (FFN-SkipLLM) & 78.89 & 7.55 & 8.11\\
    \bottomrule
    \end{tabular}
    \vspace{-0.5em}
    \caption{Performance comparison of Autoregressive Decoding with $\sim 20\%$ layers skipped using SoTA methods (SkipDecode, ShortGPT) wrt. our proposed input-adaptive FFN-SkipLLM on knowledge-intensive tasks.}
    \vspace{-1em}
    \label{tab:knowledge_evaluation_2}
\end{table}

\section{FFN-SkipLLM: A Fine-grained Input-adaptive FFN Skipping}

\subsection{Preliminaries and Motivation}
\begin{figure}
    \centering
    \includegraphics[width=0.99\linewidth]{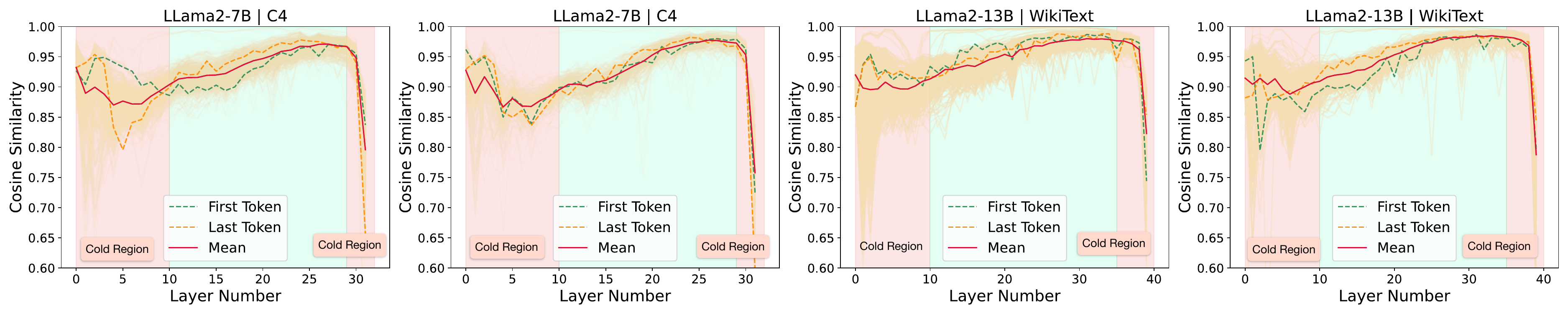}
    \vspace{-1em}
    \caption{Cosine similarity  across embedding dimension of a token tensor entering before and after the FFN block of different layers in LLaMa-2 7B and 13B model. Inputs are sampled at random from Wikitext ad C4 datasets and the mean curve indicates the average cosine similarity across 128 generated tokens. Red regions are termed \textit{cold regions} in our work and skipping FFN blocks within this region significantly hurt LLMs performance.}
    \label{fig:cosine_similarity}
    \vspace{-1.5em}
\end{figure}

Given a autoregressive large language model (LLaMa-2 in our case) $\mathbf{M_{L}}$ with \texttt{T} layers, each layer $l_i \in \texttt{L}$ consist of two major computational blocks: Multihead-Attention block ($W_q, W_k, W_v, W_o$) and FFN block ($FF_{W1}, FF_{W2}, FF_{W3}$). Table \ref{tab:parameter_count} presents the approximate parameter counts occupied by these components in layer $l_i$ indicating FFN blocks occupying around two-third of total parameter counts. In pursuit of avoiding the KV issue incurred due to entire layer-skipping, we explored the redundant computation done by FFN blocks during token-by-token generation. More specifically, given a layer $l_i$, we calculated the cosine similarity across the embedding dimension of the tensor entering the a given FFN block and exiting the block.

Figure \ref{fig:cosine_similarity} presents the layerwise mean cosine similarity of 128 generated tokens across different layers in LlaMa-2 7B and 13B models where the initial input prompt was sampled from the wikitext and C4 datasets. We are motivated by the following three observations: \circled{1} surprisingly \textbf{high cosine similarity} across the embedding dimension of the tensor entering a given FFN block and exiting it indicates the existence of redundant computation; \circled{2} \textbf{monotonically increasing cosine similarity} across middle layers (yellow region) indicating redundant computation is concentrated around middle layers in the model $\mathbf{M_{L}}$; \circled{3} \textbf{existence of two cold segments} (red region) where there exists a decreasing trend of cosine similarity indicating they significantly influence the input tensor and should be left intact during our FFN blocks skipping goal. In addition, a recent work \citep{xiao2023efficient} identified the emergence of \textit{attention sink} attributed to the strong attention scores towards initial tokens in autoregressive token by token generation. Our experiments found this observation is highly effective in stabilizing the generated tokens with FFN block-skipping and reducing repetitive tokens. FFN-SkipLLM incorporates this with a hyperparameter \texttt{warm\_up\_index} to develop a high-quality KV cache for the initial few token generation before adopting FFN skipping policy. 

\begin{wrapfigure}{R}{0.55\textwidth} 
\vspace{-1.4em}
\begin{algorithm}[H]
\small
\DontPrintSemicolon
  \SetKwFunction{FMain}{Main}
  \SetKwFunction{FSum}{Sum}
  \SetKwFunction{FSub}{\texttt{generate\_with\_skip\_model}}
  \SetKwProg{Fn}{def}{:}{}
  \KwInput{\texttt{warm\_up\_index}: int; \texttt{input\_state}: tensor; \texttt{cold\_s}: int; \texttt{cold\_e}: int; \texttt{token\_index}: int}

        \vspace{0.5em}
        
  \If{$\texttt{token\_index} \leq \texttt{warm\_up\_index}$}
  {
        \texttt{generate\_with\_full\_model(token\_index, input\_state)}
  }
  \Else{
        \texttt{generate\_with\_skip\_model(token\_index, input\_state, cold\_s, cold\_e)}
  }

    \vspace{0.5em}
  \Fn{\FSub{\texttt{token\_index, input\_state, cold\_s, cold\_e}}}{
  
        \texttt{past\_state} $\leftarrow$ \texttt{input\_state} \;

        \For{$\texttt{<0 ... cold\_s>}$}
           {
                $h$ $\leftarrow$ \texttt{past\_start} + \texttt{attention(\texttt{past\_state})} \;
                
                \texttt{past\_state} $\leftarrow h$  +  \texttt{feed\_forward($h$)}
                
           }

        \texttt{skip\_state} $\leftarrow$ \texttt{False}

        \For{$\texttt{<cold\_s ... cold\_e>}$}
           {
                $h$ $\leftarrow$ \texttt{past\_start} + \texttt{attention(\texttt{past\_state})} \;

                \If{$\texttt{skip\_state} == \texttt{False}$}
                {
                    \texttt{temp} $\leftarrow h$  +  \texttt{feed\_forward($h$)}
                    
                    \texttt{sim\_score} $\leftarrow$ \texttt{cosine ($h$, temp)}

                    \If{$\texttt{sim\_score} \geq \texttt{sim\_threshold}$}{

                        \texttt{skip\_state} $\leftarrow$ \texttt{True}
                        
                    }

                    \texttt{past\_state} $\leftarrow$ \texttt{temp}
                    
                }
                \Else{

                    \texttt{past\_state} $\leftarrow h$
                
                }
               
           }
        
        \For{$\texttt{<cold\_e ... num\_layers>}$}
           {
                $h$ $\leftarrow$ \texttt{past\_state} + \texttt{attention(\texttt{past\_state})} \;
                
                \texttt{past\_state} $\leftarrow h$  +  \texttt{feed\_forward($h$)}
                
           }
        
  }

\caption{Pseudocode for our Input-Adaptive FFN-SkipLLM}
\label{alg:ffn-skip}
\end{algorithm}
\vspace{-1em}
\end{wrapfigure}

\subsection{Methodology}

In this section, we will discuss our proposed methodology for input-adaptive FFN-SkipLLM. As discussed earlier, FFN-SkipLLM capitalizes the redundant computational cost inhibited by FFN blocks across deep autoregressive LLMs for token generation. As shown in Figure \ref{fig:cosine_similarity}, given the model  $\mathbf{M_{L}}$, its layers can be categorized into two regions: \textit{cold regions} (FFNs are non-redundant) and \textit{non-cold regions} (FFNs tend to be redundant). Cold regions (red) encompass the first few layers (\texttt{cold\_s}) and the last few layers (\texttt{cold\_e}) and they can be identified using a small calibration set from Wikitext/C4. FFN-SkipLLM uses an extra hyperparamter \texttt{warm\_up\_index} which represents how many initial first tokens will not undergo any layer-skipping to capitalize the attention sink observation. 

Algorithm \ref{alg:ffn-skip} illustrates the pseudocode for FFN-SkipLLM. A typical transformer layer performs two heavy operations: attention calculation and feed-forward transformation. Our proposed method allows both operations in cold regions but facilitates skipping feed-forward transformation within the non-cold regions. Our input adaptivity comes from tracking the cosine similarity of the token features before and after FFN blocks and deciding when to start skipping given a \texttt{sim\_thresold}. More specifically, based on our \textit{monotonically increasing} cosine similarity in non-cold regions, we \textbf{greedily} skip $k$ FFN blocks from the subsequent layers. 


\section{Experimental Results}
\textbf{Baseline Details:} To empirically evaluate the performance gains enabled by our proposed FFN-SkipLLM across multiple knowledge-intensive tasks. We aim to investigate how well FNN block skipping can retain the ability to access factoid answers ingested during pretraining, perform multi-turn instruction following, and in-context summarization. Our baselines are: \circled{1} \textit{full model} which indicate the maximum capability of LLM under consideration; \circled{2} \textit{random skip} where FFN-blocks are dropped at random without giving careful consideration of cold and non-cold regions; \circled{3} \textit{no input adaptive} where we do not track the cosine similarity per token and FFN-blocks are dropped at random from the non-cold region. Our baselines are constructed to carefully validate the effect of our observations in FFN-SkipLLM.

\subsection{Factoid-based Question Answering}
\label{sec:factoid-qa}
\textbf{Task Definition and Rationale.} Factoid-based Question Answering (Factoid-QA) \citep{Iyyer2014ANN}, which asks precise facts about entities, is a long-standing problem in NLP. A typical Factoid-QA task aims to search for entities or entity attributes from a knowledge graph, and it is widely used as a tool in academia, commercial search engines, and conversational assistants. Modern LLMs are trained on gigantic text corpora ingesting a large amount of world knowledge about entities and their relationships during pre-training, and have unique abilities to generate factually correct responses to user queries. In this task setting, we aim to investigate \textit{how our input-adaptive FFN block skipping impacts LLMs' ability to answer natural language questions using facts,  i.e., entities or attributes knowledge ingested within them during pre-training?}

\begin{table}[]
\small
    \centering
    \begin{tabular}{c|cccc}
    \toprule
        \textbf{Method} & \textbf{$\sim$5\%} & \textbf{$\sim$15\%} & \textbf{$\sim$25\%}  & \textbf{$\sim$35\%}\\
    \midrule
        \rowcolor[gray]{0.9}
        Full Model & \multicolumn{4}{c}{79.02\%}\\
        Baseline 1 (Random Skip) & 77.32\% & 72.96\% & 49.22\% & 31.07\%\\
        Baseline 2 (No input adaptive) & 78.92\% & 77.71\% & 74.13\% & 69.93\%\\
    \midrule
        \rowcolor[gray]{0.9}
        Ours (FFN-SkipLLM) & 80.05\% & 78.42\% & 78.09\% & 75.61\%\\
    \bottomrule
    \end{tabular}
    \caption{Performance comparison of our baselines with varying layer skip ratios wrt. proposed input-adaptive FFN-SkipLLM on Factoid-based QA.}
    \label{tab:knowledge_evaluation}
    \vspace{-1em}
\end{table}

\begin{center}
\scriptsize
\begin{tcolorbox}[width=0.95\textwidth]
\paragraph{\textcolor{cyan}{\textbf{Prompt Design:}}} Please give answer to this question: $<$\texttt{QUESTION}$>$ The answer is  
\vspace{0.1cm}
\paragraph{\textcolor{blue}{\textbf{Example:}}} Please give answer to this question: \texttt{Who is the prime minister of India?}  
\vspace{0.1cm}
\paragraph{\textcolor{violet}{\textbf{Model Response:}}} Please give answer to this question: \texttt{Who is the prime minister of India?} The current Prime Minister of India is \texttt{\textcolor{teal}{Narendra Modi}}. He has been serving as the Prime
Minister since 2014. The Prime Minister of India is a political leader who is elected by the
parliamentarians and serves as the head of the government. The Prime Minister is
responsible for implementing policies, programs, and initiatives to improve the economic,
social, and political well-being of the country. 
\end{tcolorbox}
\end{center}

\textbf{Dataset Details and Results.} We use FreebaseQA \citep{Jiang2019FreebaseQAAN} which is a dataset for open-domain QA over the Freebase knowledge graph. The QA pairs are collected from various sources, including the TriviaQA dataset  \citep{joshi2017triviaqa} and other trivia websites (QuizBalls, QuizZone, KnowQuiz), and are matched against Freebase to generate relevant subject-predicate-object triples that were further verified by human annotators. TriviaQA dataset shows rich linguistic variation and complexity, making it a good testbed for evaluating knowledge ingested within LLMs. 

The results of various baseline methods and FFN-SkipLLM are demonstrated in Table \ref{tab:knowledge_evaluation}. It is interesting to observe that FFN-SkipLLM with $\sim$5\% skip ratio per token can outperform the full model performance. A careful study of Baselines 1 and 2 indicates the effectiveness of our observation of cold vs non-cold regions for FFN-block skipping. Note that at a high skip ratio, the performance of the random baseline is significantly worse with $\geq$50\% performance drop. On the other hand, we can also note that our input-adaptive FFN-SkipLLM is highly robust in retaining a large fraction of full model performance in comparison to Baseline 2.


\begin{figure}[h]
    \centering
    \includegraphics[width=0.99\linewidth]{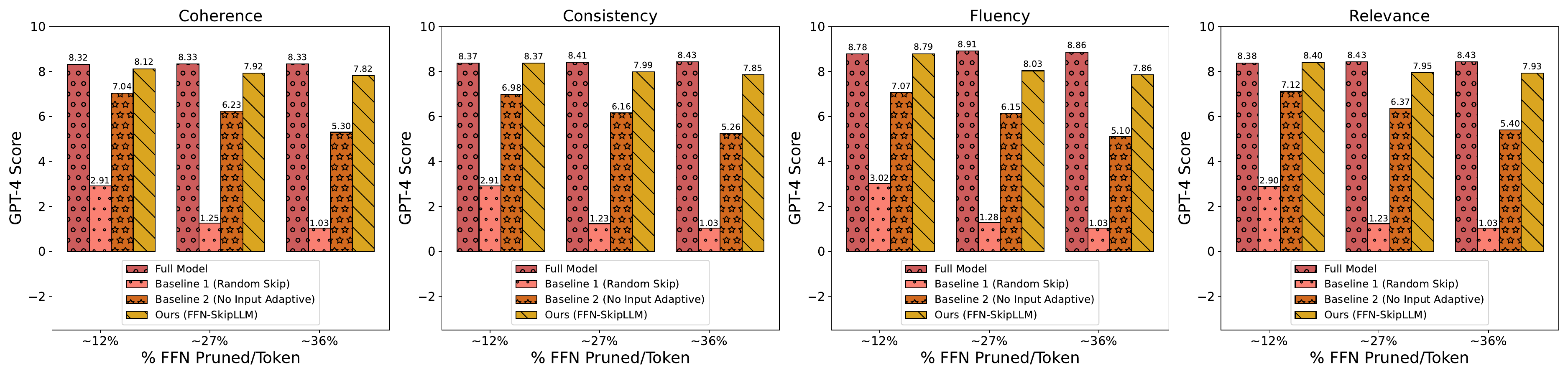}
    \includegraphics[width=0.99\linewidth]{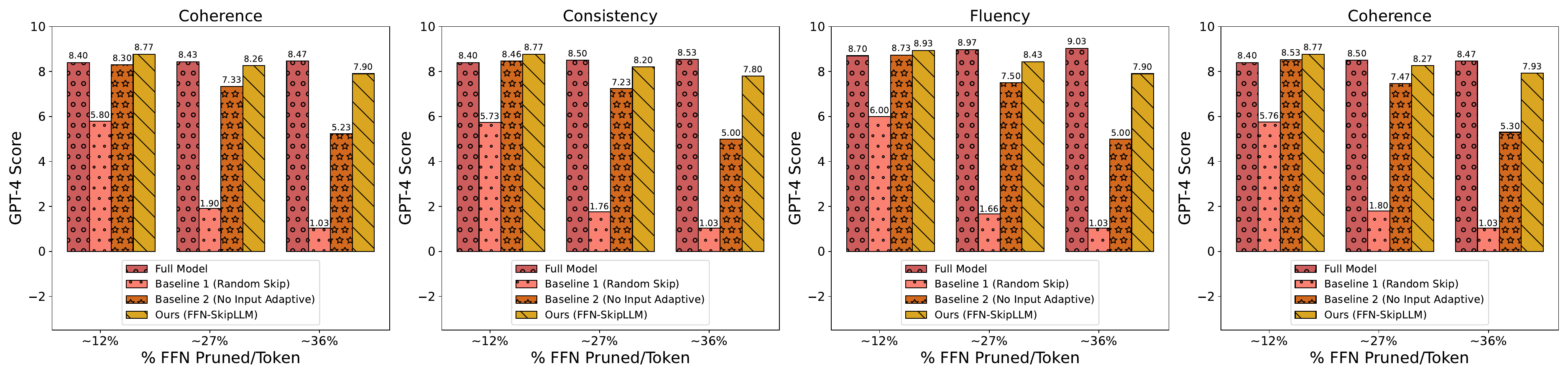}
    \includegraphics[width=0.99\linewidth]{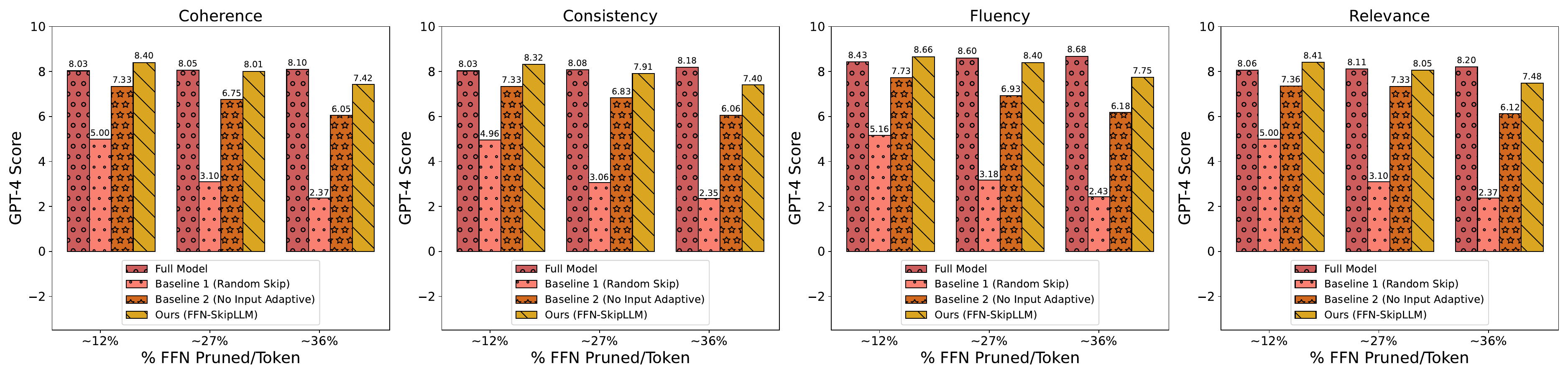}
    \vspace{-1em}
    \caption{Performance comparison of our baselines wrt. FFN-SkipLLM for in-context summarization of small (row 1), medium (row 2), and large (row 3) stories while preserving coherence, consistency, fluency, and relevance.}
    \label{fig:text_summarization_evaluation}
    \vspace{-1em}
\end{figure}


\subsection{In-context Variable Length Text Summarization}
\label{sec:summarization}
\textbf{Task Formulation and Details.} 
Modern LLMs have shown astonishing success in summarizing long-context documents in both abstractive and extractive settings. However, it is \textbf{yet not explored} how FFN block skipping impacts LLMs' capability for summarization. In this task setting, we aim to investigate \textit{how well autoregressive decoding with FFN block skipping hold onto consistency, coherence, fluency, and relevance when prompted to summarize textual information of varying length (small, medium, and large) in abstractive setting} \citep{jain2023multi}. For evaluation, 
similar to~\cite{zheng2023judging}, we propose to use GPT-4 as a judge, which compares the compressed LLM generated summaries wrt. GPT-3.5 (text-davinci-003) generated summaries.

\begin{center}
\vspace{-1em}
\scriptsize
\begin{tcolorbox}[width=0.99\textwidth]
\paragraph{\textcolor{cyan}{\textbf{Prompt Design:}}} A chat between a curious user and an artificial intelligence assistant. The assistant gives helpful, detailed, and polite answers to the user's questions. USER: \texttt{Summarize the given story in less than 150 words while preserving high coherence, consistency, fluency, and relevance.}$\backslash$n$\backslash$n $<$\texttt{STORY}$>$. 

ASSISTANT:
\vspace{0.1cm}
\paragraph{\textcolor{blue}{\textbf{Example:}}} A chat between a curious user and an artificial intelligence assistant. The assistant gives helpful, detailed, and polite answers to the user's questions. USER: \texttt{Summarize the given story in less than 150 words while preserving high coherence, consistency, fluency, and relevance.}$\backslash$n$\backslash$n\texttt{Libyan and U.S. officials say the two governments held face-to-face talks in Tunisia ...have denied previous reports of talks with the government}. 

ASSISTANT:

\end{tcolorbox}
\vspace{-0.5em}
\end{center}

\textbf{Dataset Details and Results} We use a popular summarization dataset CNN/DailyMail \citep{chen2016thorough} for evaluation, which is an English-language dataset containing just over 300k unique news articles written by journalists at CNN and DailyMail. We created 3 subset categories \{small ($\leq$470 words), medium ($\geq$470 and $\leq$ 790 words), and large ($\geq$ 790 words)\} of stories, each with 100 articles reflecting word distribution of CNN/DailyMail to minimize OpenAI API costs. 

Figure \ref{fig:text_summarization_evaluation} summarizes the result of the variable length text summarization task. One interesting observation we find is that with increasing in-context stories for summarization, we found that the performance of random baseline improves. Upon digging we found that it start copying random text snippets from the in-context story directly into the summary which led to a comparatively better GPT-4 evaluation score. With an increasing skip ratio, we found that the performance gap between FFN-SkipLLM and our baselines increases. Moreover, at $\sim$10-12\% skip ratio we found that GPT-4 consistently ranks our summary better than the full model across coherence, consistency, fluency, and relevance.

\subsection{Mutlti-turn Conversation and Instruction Following}
\label{sec:mt-bench}
\textbf{Task Formulation and Rationale. } In this task setting, we investigate \textit{how FFN block skipping impacts the LLMs' ability to answer open-ended questions and evaluate their multi-turn conversational and instruction-following ability – two critical elements for human preference}. Evaluating AI chatbots is a challenging task, as it requires examining language understanding, reasoning, and context awareness. To compare the performance of compressed LLMs' responses, we closely follow the prompt design setting in MT-Bench \citep{zheng2023judging} using GPT-4 as a judge. We prompt GPT-4 to rate the answers generated by compressed LLMs wrt. GPT-3.5 (text-davinci-003) model based on varying metrics (\emph{e.g.}, correctness, helpfulness, logic, accuracy, \emph{etc.}) on a scale of \texttt{[0-10]} with detailed explanations.

\begin{figure}
    \centering
    \includegraphics[width=\linewidth]{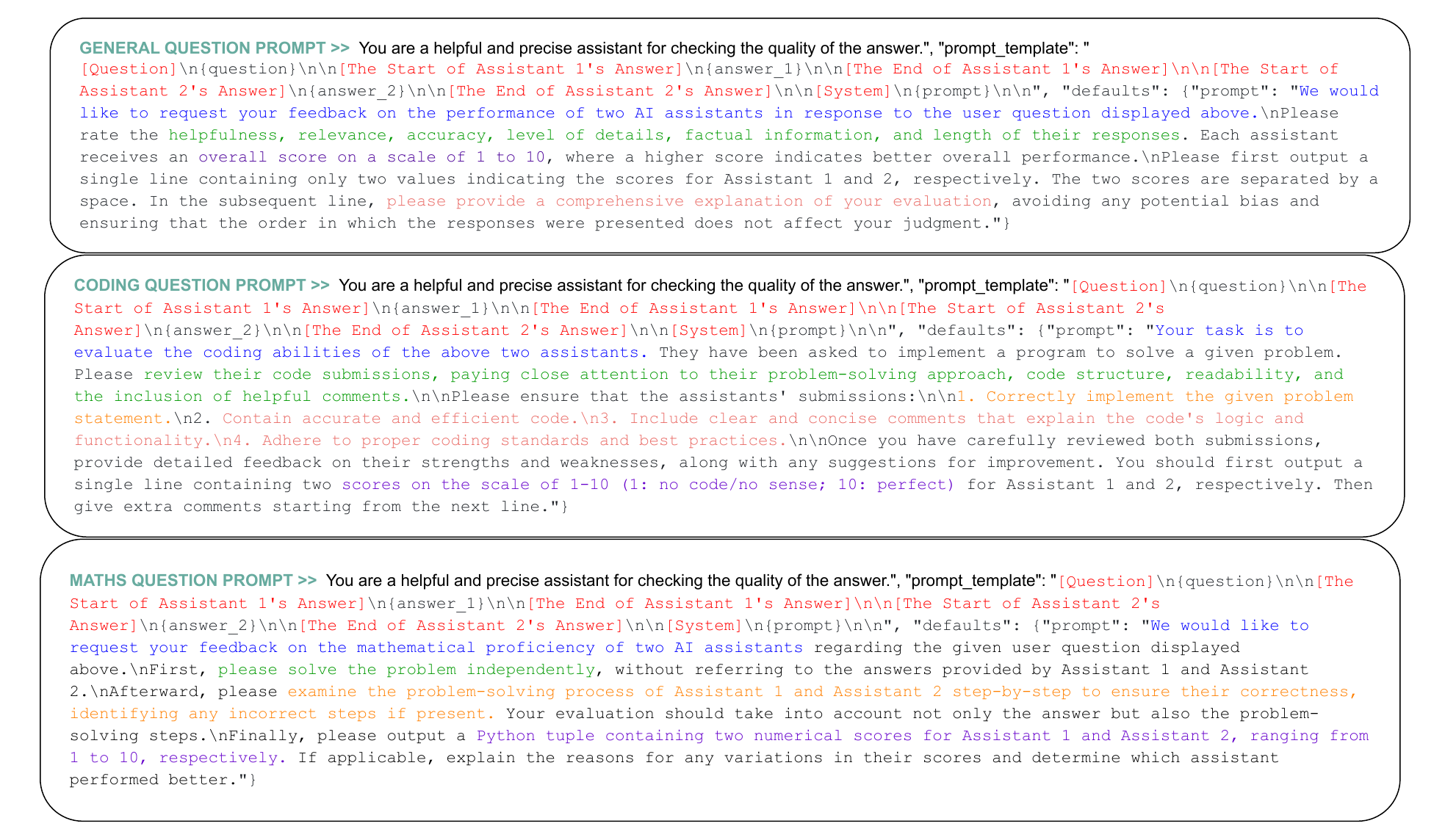}
    \vspace{-2.2em}
    \caption{Examples of prompts used for different categories to evaluate the compressed LLM ASSISTANT \emph{wrt.} GPT-3.5 ASSISTANT using GPT-4 as a Judge.}
    \label{fig:multi_turn_judge}
    \vspace{-0.5em}
\end{figure}

\begin{figure}
    \centering
    \includegraphics[width=0.99\linewidth]{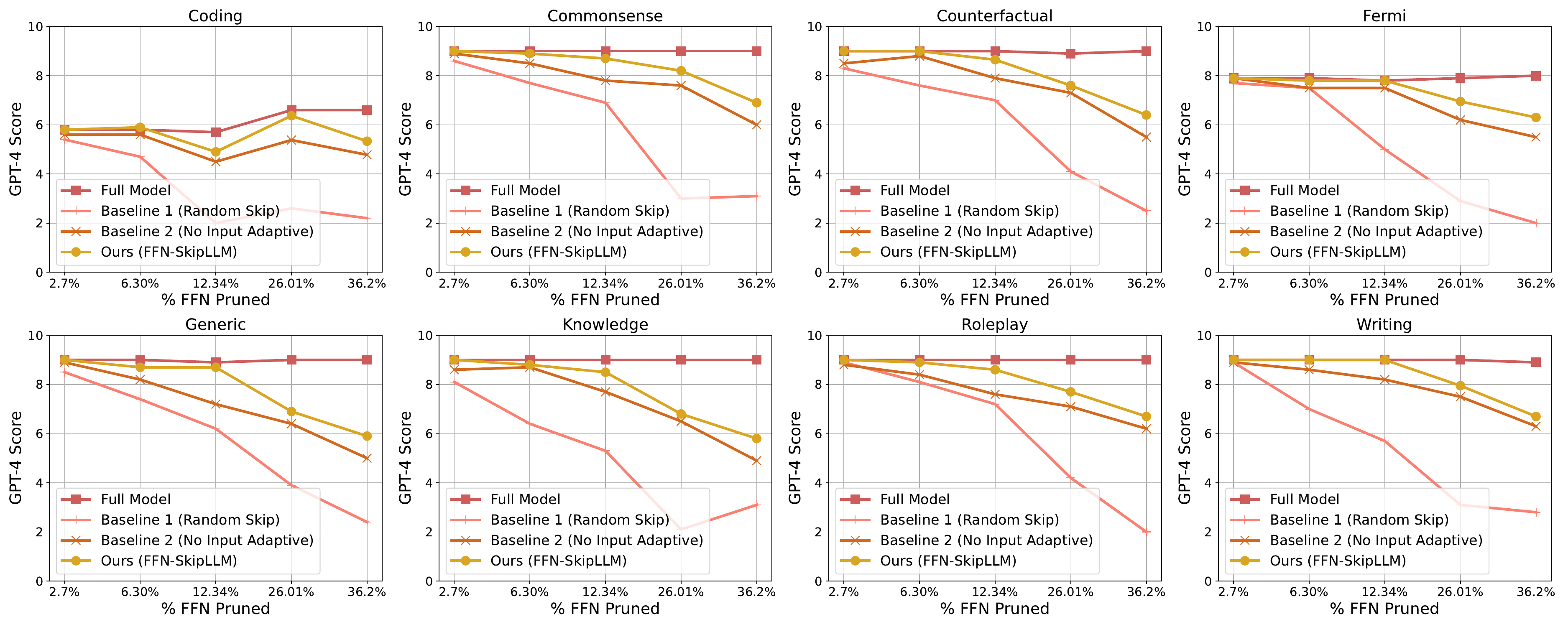}
    \vspace{-1em}
    \caption{Performance comparison of our baselines with varying layer skip ratios wrt.  FFN-SkipLLM on multi-turn conversation across 8 different categories. }
    \label{fig:mt_bench}
    \vspace{-1.5em}
\end{figure}

\begin{center}
\scriptsize
\begin{tcolorbox}[width=0.995\textwidth]
\paragraph{\textcolor{cyan}{\textbf{Prompt Design:}}} A chat between a curious user and an artificial intelligence assistant. The assistant gives helpful, detailed, and polite answers to the user's questions. USER: $<$\texttt{QUESTION}$>$ 

ASSISTANT:
\vspace{0.1cm}
\paragraph{\textcolor{blue}{\textbf{Example:}}} A chat between a curious user and an artificial intelligence assistant. The assistant gives helpful, detailed, and polite answers to the user's questions. USER: \texttt{How can I improve my time management skills?}

ASSISTANT:
\vspace{0.1cm}
\end{tcolorbox}
\end{center}

\textbf{Dataset Details and Results.} We rely on the 80 high-quality multi-turn questions identified in MT-Bench \citep{zheng2023judging}. This setting covers common-use human-centric interaction with LLMs, and focuses on challenging questions to differentiate models. We used 8 common categories of user prompts to guide the prompt construction to interact with compressed LLMs: writing, roleplay, extraction, reasoning, math, coding, \emph{etc}. For each category, we adopted manually designed 10 multi-turn questions from MT-Bench to evaluate our compressed models. 

Figure \ref{fig:mt_bench} presents the performance comparison of our baseline models across 8 different categories. It is surprising to observe that across some categories such as coding, fermi, and commonsense; FFN-SkipLLM perform quite match the performance of the full model comfortably up to $\sim$25\%  skip ratio per token. Unlike identified by \citep{men2024shortgpt} that layer dropping fails on generative tasks, it is important to acknowledge our careful FFN block dropping can significantly reduce hallucination across knowledge-intensive tasks. Note that our random skip baseline observes a terminal decline in performance even with a slop rato of 10-15\% which suggests the importance of cold regions and input-adaptivity.

\section{Background Work}

Recent advances in model compression (pruning, quantization, and distillation) have been very successful in democratizing LLMs, allowing them to perform inference on consumer-grade GPUs. In contrast to their static nature, input-dependent early-exit or layer-dropping strategies present a unique potential for faster inference for new gigantic auto-regressive models during token-by-token generation. The majority of existing approaches primarily has been around BERT-scale encoder models ~\citep{dynabert, cascadebert, fastbert, deebert, zhu-2021-leebert}. A notable challenge in auto-regressive generation tasks is managing Key-Value (KV) caching, a process that stores the keys and values from attention layers corresponding to previously generated tokens to accelerate sequence generation. However, if a token is generated via early exiting, the KV caches for all subsequent layers are missing, complicating the generation of future tokens that exit at layers beyond the initial exiting layer. This challenge has been acknowledged in the literature, and various strategies have been proposed to address it. One method~\citep{elbayad2019depth, li2021accelerating, calm} duplicates the hidden states from the current token's exiting layer to subsequent layers, which act as the KV cache for generating future tokens. Although being efficient, it causes deviation in the inference process and generates sub-optimal outputs. Another approach~\citep{del2023skipdecode} pre-determines the exiting layers for all tokens, which guarantees later tokens always exits at earlier layers, thus ensuring KV caches are always present. However,this approach suffers from degrading performance for as token length increases, and pinpointing the optimal exiting parameters to balance model performance with inference efficiency is non-trivial. The third strategy~\citep{bae-etal-2023-fast, tang2023deed} stores the hidden states of previous tokens that early-exited. When a KV cache missing occurs, a batched forward pass wtih current and recent tokens is conducted, materializing the missing KV cache. In the worst-case scenario, this approach requires utilizing the full network, thus negating the intended efficiency benefits. In contrast to these work, our work explores an orthogonal direction to layer skipping and focuses on FFN-block skipping which circumvents the hassle and issues with KV caching and can effectively ignore two-thirds of parameter counts.

\section{Conclusion and Limitations}
In this paper, we explore an orthogonal dimension for layer-skipping and early-exit strategies that suffer from KV cache issues leading to the hallucination of misinformation and token collapse. We propose \textbf{FFN-SkipLLM},  a novel fine-grained skip strategy of autoregressive LLMs which is an input-adaptive feed-forward skipping strategy that can skip $\sim$ 25-30\% of FFN blocks of LLMs
with marginal change in performance on knowledge-intensive tasks. FNN-Skip LLM is built on the core observation of monotonically increasing redundancy within the FFN blocks of LLMs. One major limitation of our work is scaling FFN-SkipLLM for non-trivial skipping ratios ($\geq$ 35\%) without a significant performance drop. Our future work includes exploring parameter-efficient fine-tuning techniques to push the performance of high skip ratios. Note that FFN-SkipLLM can be easily combined to with modern advancements in sparsity and quantization for favourable speedups.

\bibliography{colm2024_conference}

\begin{thebibliography}{64}
\providecommand{\natexlab}[1]{#1}
\providecommand{\url}[1]{\texttt{#1}}
\expandafter\ifx\csname urlstyle\endcsname\relax
  \providecommand{\doi}[1]{doi: #1}\else
  \providecommand{\doi}{doi: \begingroup \urlstyle{rm}\Url}\fi

\bibitem[Bae et~al.(2023{\natexlab{a}})Bae, Ko, Song, and Yun]{bae-etal-2023-fast}
Sangmin Bae, Jongwoo Ko, Hwanjun Song, and Se-Young Yun.
\newblock Fast and robust early-exiting framework for autoregressive language models with synchronized parallel decoding.
\newblock In Houda Bouamor, Juan Pino, and Kalika Bali (eds.), \emph{Proceedings of the 2023 Conference on Empirical Methods in Natural Language Processing}, pp.\  5910--5924, Singapore, December 2023{\natexlab{a}}. Association for Computational Linguistics.
\newblock \doi{10.18653/v1/2023.emnlp-main.362}.
\newblock URL \url{https://aclanthology.org/2023.emnlp-main.362}.

\bibitem[Bae et~al.(2023{\natexlab{b}})Bae, Ko, Song, and Yun]{bae2023fast}
Sangmin Bae, Jongwoo Ko, Hwanjun Song, and Se-Young Yun.
\newblock Fast and robust early-exiting framework for autoregressive language models with synchronized parallel decoding.
\newblock \emph{arXiv preprint arXiv:2310.05424}, 2023{\natexlab{b}}.

\bibitem[Bahdanau et~al.(2014)Bahdanau, Cho, and Bengio]{bahdanau2014neural}
Dzmitry Bahdanau, Kyunghyun Cho, and Yoshua Bengio.
\newblock Neural machine translation by jointly learning to align and translate.
\newblock \emph{arXiv preprint arXiv:1409.0473}, 2014.

\bibitem[Chen et~al.(2016)Chen, Bolton, and Manning]{chen2016thorough}
Danqi Chen, Jason Bolton, and Christopher~D Manning.
\newblock A thorough examination of the cnn/daily mail reading comprehension task.
\newblock \emph{arXiv preprint arXiv:1606.02858}, 2016.

\bibitem[Chen et~al.(2024)Chen, Zhao, Jaiswal, Shah, and Wang]{chen2024llaga}
Runjin Chen, Tong Zhao, Ajay Jaiswal, Neil Shah, and Zhangyang Wang.
\newblock Llaga: Large language and graph assistant.
\newblock \emph{arXiv preprint arXiv:2402.08170}, 2024.

\bibitem[Chen et~al.(2020)Chen, Frankle, Chang, Liu, Zhang, Wang, and Carbin]{chen2020lottery}
Tianlong Chen, Jonathan Frankle, Shiyu Chang, Sijia Liu, Yang Zhang, Zhangyang Wang, and Michael Carbin.
\newblock The lottery ticket hypothesis for pre-trained bert networks.
\newblock \emph{Advances in neural information processing systems}, 33:\penalty0 15834--15846, 2020.

\bibitem[Chen et~al.(2023{\natexlab{a}})Chen, Pan, Li, Ding, and Zhou]{Chen2023EELLMLT}
Yanxi Chen, Xuchen Pan, Yaliang Li, Bolin Ding, and Jingren Zhou.
\newblock Ee-llm: Large-scale training and inference of early-exit large language models with 3d parallelism.
\newblock \emph{ArXiv}, abs/2312.04916, 2023{\natexlab{a}}.
\newblock URL \url{https://api.semanticscholar.org/CorpusID:266149909}.

\bibitem[Chen et~al.(2023{\natexlab{b}})Chen, Pan, Li, Ding, and Zhou]{chen2023ee}
Yanxi Chen, Xuchen Pan, Yaliang Li, Bolin Ding, and Jingren Zhou.
\newblock Ee-llm: Large-scale training and inference of early-exit large language models with 3d parallelism.
\newblock \emph{arXiv preprint arXiv:2312.04916}, 2023{\natexlab{b}}.

\bibitem[Chen et~al.(2023{\natexlab{c}})Chen, Mao, Li, Jin, Wen, Wei, Wang, Yin, Fan, Liu, et~al.]{chen2023exploring}
Zhikai Chen, Haitao Mao, Hang Li, Wei Jin, Hongzhi Wen, Xiaochi Wei, Shuaiqiang Wang, Dawei Yin, Wenqi Fan, Hui Liu, et~al.
\newblock Exploring the potential of large language models (llms) in learning on graphs.
\newblock \emph{arXiv preprint arXiv:2307.03393}, 2023{\natexlab{c}}.

\bibitem[Del~Corro et~al.(2023)Del~Corro, Del~Giorno, Agarwal, Yu, Awadallah, and Mukherjee]{del2023skipdecode}
Luciano Del~Corro, Allie Del~Giorno, Sahaj Agarwal, Bin Yu, Ahmed Awadallah, and Subhabrata Mukherjee.
\newblock Skipdecode: Autoregressive skip decoding with batching and caching for efficient llm inference.
\newblock \emph{arXiv preprint arXiv:2307.02628}, 2023.

\bibitem[Dettmers et~al.(2023{\natexlab{a}})Dettmers, Pagnoni, Holtzman, and Zettlemoyer]{Dettmers2023QLoRAEF}
Tim Dettmers, Artidoro Pagnoni, Ari Holtzman, and Luke Zettlemoyer.
\newblock Qlora: Efficient finetuning of quantized llms.
\newblock \emph{ArXiv}, abs/2305.14314, 2023{\natexlab{a}}.
\newblock URL \url{https://api.semanticscholar.org/CorpusID:258841328}.

\bibitem[Dettmers et~al.(2023{\natexlab{b}})Dettmers, Svirschevski, Egiazarian, Kuznedelev, Frantar, Ashkboos, Borzunov, Hoefler, and Alistarh]{Dettmers2023SpQRAS}
Tim Dettmers, Ruslan Svirschevski, Vage Egiazarian, Denis Kuznedelev, Elias Frantar, Saleh Ashkboos, Alexander Borzunov, Torsten Hoefler, and Dan Alistarh.
\newblock Spqr: A sparse-quantized representation for near-lossless llm weight compression.
\newblock \emph{ArXiv}, abs/2306.03078, 2023{\natexlab{b}}.
\newblock URL \url{https://api.semanticscholar.org/CorpusID:259076379}.

\bibitem[Duan et~al.(2023)Duan, Liu, Chua, Yan, Ooi, Xie, and He]{duan2023simteg}
Keyu Duan, Qian Liu, Tat-Seng Chua, Shuicheng Yan, Wei~Tsang Ooi, Qizhe Xie, and Junxian He.
\newblock Simteg: A frustratingly simple approach improves textual graph learning.
\newblock \emph{arXiv preprint arXiv:2308.02565}, 2023.

\bibitem[Elbayad et~al.(2019)Elbayad, Gu, Grave, and Auli]{elbayad2019depth}
Maha Elbayad, Jiatao Gu, Edouard Grave, and Michael Auli.
\newblock Depth-adaptive transformer.
\newblock \emph{arXiv preprint arXiv:1910.10073}, 2019.

\bibitem[Fabbri et~al.(2019)Fabbri, Li, She, Li, and Radev]{fabbri2019multi}
Alexander~R Fabbri, Irene Li, Tianwei She, Suyi Li, and Dragomir~R Radev.
\newblock Multi-news: A large-scale multi-document summarization dataset and abstractive hierarchical model.
\newblock \emph{arXiv preprint arXiv:1906.01749}, 2019.

\bibitem[Frankle \& Carbin(2019)Frankle and Carbin]{frankle2018the}
Jonathan Frankle and Michael Carbin.
\newblock The lottery ticket hypothesis: Finding sparse, trainable neural networks.
\newblock In \emph{International Conference on Learning Representations}, 2019.
\newblock URL \url{https://openreview.net/forum?id=rJl-b3RcF7}.

\bibitem[Frantar et~al.(2022)Frantar, Ashkboos, Hoefler, and Alistarh]{Frantar2022GPTQAP}
Elias Frantar, Saleh Ashkboos, Torsten Hoefler, and Dan Alistarh.
\newblock Gptq: Accurate post-training quantization for generative pre-trained transformers.
\newblock \emph{ArXiv}, abs/2210.17323, 2022.
\newblock URL \url{https://api.semanticscholar.org/CorpusID:253237200}.

\bibitem[Hou et~al.(2020)Hou, Huang, Shang, Jiang, Chen, and Liu]{dynabert}
Lu~Hou, Zhiqi Huang, Lifeng Shang, Xin Jiang, Xiao Chen, and Qun Liu.
\newblock Dynabert: dynamic bert with adaptive width and depth.
\newblock In \emph{Proceedings of the 34th International Conference on Neural Information Processing Systems}, NIPS'20, Red Hook, NY, USA, 2020. Curran Associates Inc.
\newblock ISBN 9781713829546.

\bibitem[Iyyer et~al.(2014)Iyyer, Boyd-Graber, Claudino, Socher, and Daum{\'e}]{Iyyer2014ANN}
Mohit Iyyer, Jordan~L. Boyd-Graber, Leonardo Max~Batista Claudino, Richard Socher, and Hal Daum{\'e}.
\newblock A neural network for factoid question answering over paragraphs.
\newblock In \emph{Conference on Empirical Methods in Natural Language Processing}, 2014.
\newblock URL \url{https://api.semanticscholar.org/CorpusID:216034672}.

\bibitem[Jain et~al.(2023)Jain, Keshava, Sathyendra, Fernandes, Liu, Neubig, and Zhou]{jain2023multi}
Sameer Jain, Vaishakh Keshava, Swarnashree~Mysore Sathyendra, Patrick Fernandes, Pengfei Liu, Graham Neubig, and Chunting Zhou.
\newblock Multi-dimensional evaluation of text summarization with in-context learning.
\newblock \emph{arXiv preprint arXiv:2306.01200}, 2023.

\bibitem[Jaiswal et~al.(2021)Jaiswal, Tang, Ghosh, Rousseau, Peng, and Ding]{jaiswal2021radbert}
Ajay Jaiswal, Liyan Tang, Meheli Ghosh, Justin~F Rousseau, Yifan Peng, and Ying Ding.
\newblock Radbert-cl: Factually-aware contrastive learning for radiology report classification.
\newblock In \emph{Machine Learning for Health}, pp.\  196--208. PMLR, 2021.

\bibitem[Jaiswal et~al.(2023{\natexlab{a}})Jaiswal, Liu, Chen, and Wang]{jaiswal2023emergence}
Ajay Jaiswal, Shiwei Liu, Tianlong Chen, and Zhangyang Wang.
\newblock The emergence of essential sparsity in large pre-trained models: The weights that matter.
\newblock \emph{arXiv preprint arXiv:2306.03805}, 2023{\natexlab{a}}.

\bibitem[Jaiswal et~al.(2022)Jaiswal, Ma, Chen, Ding, and Wang]{jaiswal2022training}
Ajay~Kumar Jaiswal, Haoyu Ma, Tianlong Chen, Ying Ding, and Zhangyang Wang.
\newblock Training your sparse neural network better with any mask.
\newblock In \emph{International Conference on Machine Learning}, pp.\  9833--9844. PMLR, 2022.

\bibitem[Jaiswal et~al.(2023{\natexlab{b}})Jaiswal, Liu, Chen, Ding, and Wang]{jaiswal2023instant}
Ajay~Kumar Jaiswal, Shiwei Liu, Tianlong Chen, Ying Ding, and Zhangyang Wang.
\newblock Instant soup: Cheap pruning ensembles in a single pass can draw lottery tickets from large models.
\newblock In \emph{International Conference on Machine Learning}, pp.\  14691--14701. PMLR, 2023{\natexlab{b}}.

\bibitem[Jiang et~al.(2019)Jiang, Wu, and Jiang]{Jiang2019FreebaseQAAN}
Kelvin Jiang, Dekun Wu, and Hui Jiang.
\newblock Freebaseqa: A new factoid qa data set matching trivia-style question-answer pairs with freebase.
\newblock In \emph{North American Chapter of the Association for Computational Linguistics}, 2019.
\newblock URL \url{https://api.semanticscholar.org/CorpusID:174800890}.

\bibitem[Joshi et~al.(2017)Joshi, Choi, Weld, and Zettlemoyer]{joshi2017triviaqa}
Mandar Joshi, Eunsol Choi, Daniel~S Weld, and Luke Zettlemoyer.
\newblock Triviaqa: A large scale distantly supervised challenge dataset for reading comprehension.
\newblock \emph{arXiv preprint arXiv:1705.03551}, 2017.

\bibitem[Kim et~al.(2024)Kim, Kim, Kim, Castells, Choi, Shin, and Song]{kim2024shortened}
Bo-Kyeong Kim, Geonmin Kim, Tae-Ho Kim, Thibault Castells, Shinkook Choi, Junho Shin, and Hyoung-Kyu Song.
\newblock Shortened llama: A simple depth pruning for large language models.
\newblock \emph{arXiv preprint arXiv:2402.02834}, 2024.

\bibitem[Kim et~al.(2023)Kim, Lee, Kim, Park, Yoo, Kwon, and Lee]{Kim2023MemoryEfficientFO}
Jeonghoon Kim, Jung~Hyun Lee, Sungdong Kim, Joonsuk Park, Kang~Min Yoo, Se~Jung Kwon, and Dongsoo Lee.
\newblock Memory-efficient fine-tuning of compressed large language models via sub-4-bit integer quantization.
\newblock \emph{ArXiv}, abs/2305.14152, 2023.
\newblock URL \url{https://api.semanticscholar.org/CorpusID:258841104}.

\bibitem[Lai et~al.(2023)Lai, Tian, Chen, Li, Yuan, Liu, and Jia]{lai2023lisa}
Xin Lai, Zhuotao Tian, Yukang Chen, Yanwei Li, Yuhui Yuan, Shu Liu, and Jiaya Jia.
\newblock Lisa: Reasoning segmentation via large language model.
\newblock \emph{arXiv preprint arXiv:2308.00692}, 2023.

\bibitem[Lee et~al.(2019)Lee, Ajanthan, and Torr]{lee2018snip}
Namhoon Lee, Thalaiyasingam Ajanthan, and Philip Torr.
\newblock Snip: Single-shot network pruning based on connection sensitivity.
\newblock In \emph{International Conference on Learning Representations}, 2019.
\newblock URL \url{https://openreview.net/forum?id=B1VZqjAcYX}.

\bibitem[Lee et~al.(2023)Lee, An, and Thorne]{Lee2023CanLL}
Noah Lee, Na~Min An, and James Thorne.
\newblock Can large language models infer and disagree like humans?
\newblock \emph{ArXiv}, abs/2305.13788, 2023.
\newblock URL \url{https://api.semanticscholar.org/CorpusID:258841424}.

\bibitem[Li et~al.(2021{\natexlab{a}})Li, Lin, Chen, Ren, Li, Zhou, and Sun]{cascadebert}
Lei Li, Yankai Lin, Deli Chen, Shuhuai Ren, Peng Li, Jie Zhou, and Xu~Sun.
\newblock {C}ascade{BERT}: Accelerating inference of pre-trained language models via calibrated complete models cascade.
\newblock In Marie-Francine Moens, Xuanjing Huang, Lucia Specia, and Scott Wen-tau Yih (eds.), \emph{Findings of the Association for Computational Linguistics: EMNLP 2021}, pp.\  475--486, Punta Cana, Dominican Republic, November 2021{\natexlab{a}}. Association for Computational Linguistics.
\newblock \doi{10.18653/v1/2021.findings-emnlp.43}.
\newblock URL \url{https://aclanthology.org/2021.findings-emnlp.43}.

\bibitem[Li et~al.(2024)Li, Shetty, Kamath, Jaiswal, Jiang, Ding, and Kim]{li2024cancergpt}
Tianhao Li, Sandesh Shetty, Advaith Kamath, Ajay Jaiswal, Xiaoqian Jiang, Ying Ding, and Yejin Kim.
\newblock Cancergpt for few shot drug pair synergy prediction using large pretrained language models.
\newblock \emph{npj Digital Medicine}, 7\penalty0 (1):\penalty0 40, 2024.

\bibitem[Li et~al.(2021{\natexlab{b}})Li, Shao, Sun, Yan, Qiu, and Huang]{li2021accelerating}
Xiaonan Li, Yunfan Shao, Tianxiang Sun, Hang Yan, Xipeng Qiu, and Xuanjing Huang.
\newblock Accelerating bert inference for sequence labeling via early-exit.
\newblock \emph{arXiv preprint arXiv:2105.13878}, 2021{\natexlab{b}}.

\bibitem[Lian et~al.(2023)Lian, Li, Yala, and Darrell]{lian2023llm}
Long Lian, Boyi Li, Adam Yala, and Trevor Darrell.
\newblock Llm-grounded diffusion: Enhancing prompt understanding of text-to-image diffusion models with large language models.
\newblock \emph{arXiv preprint arXiv:2305.13655}, 2023.

\bibitem[Lin et~al.(2023)Lin, Tang, Tang, Yang, Dang, and Han]{Lin2023AWQAW}
Ji~Lin, Jiaming Tang, Haotian Tang, Shang Yang, Xingyu Dang, and Song Han.
\newblock Awq: Activation-aware weight quantization for llm compression and acceleration.
\newblock \emph{ArXiv}, abs/2306.00978, 2023.
\newblock URL \url{https://api.semanticscholar.org/CorpusID:258999941}.

\bibitem[Liu et~al.(2023{\natexlab{a}})Liu, Liu, Zhou, Ye, Chong, Zhou, Xie, Cao, Wang, You, et~al.]{liu2023llmrec}
Junling Liu, Chao Liu, Peilin Zhou, Qichen Ye, Dading Chong, Kang Zhou, Yueqi Xie, Yuwei Cao, Shoujin Wang, Chenyu You, et~al.
\newblock Llmrec: Benchmarking large language models on recommendation task.
\newblock \emph{arXiv preprint arXiv:2308.12241}, 2023{\natexlab{a}}.

\bibitem[Liu et~al.(2023{\natexlab{b}})Liu, Chen, Zhang, Chen, Huang, Jaiswal, and Wang]{liu2023sparsity}
Shiwei Liu, Tianlong Chen, Zhenyu Zhang, Xuxi Chen, Tianjin Huang, Ajay Jaiswal, and Zhangyang Wang.
\newblock Sparsity may cry: Let us fail (current) sparse neural networks together!
\newblock \emph{arXiv preprint arXiv:2303.02141}, 2023{\natexlab{b}}.

\bibitem[Liu et~al.(2020)Liu, Zhou, Wang, Zhao, Deng, and Ju]{fastbert}
Weijie Liu, Peng Zhou, Zhiruo Wang, Zhe Zhao, Haotang Deng, and Qi~Ju.
\newblock {F}ast{BERT}: a self-distilling {BERT} with adaptive inference time.
\newblock In Dan Jurafsky, Joyce Chai, Natalie Schluter, and Joel Tetreault (eds.), \emph{Proceedings of the 58th Annual Meeting of the Association for Computational Linguistics}, pp.\  6035--6044, Online, July 2020. Association for Computational Linguistics.
\newblock \doi{10.18653/v1/2020.acl-main.537}.
\newblock URL \url{https://aclanthology.org/2020.acl-main.537}.

\bibitem[Liu et~al.(2023{\natexlab{c}})Liu, Oguz, Zhao, Chang, Stock, Mehdad, Shi, Krishnamoorthi, and Chandra]{liu2023llm}
Zechun Liu, Barlas Oguz, Changsheng Zhao, Ernie Chang, Pierre Stock, Yashar Mehdad, Yangyang Shi, Raghuraman Krishnamoorthi, and Vikas Chandra.
\newblock Llm-qat: Data-free quantization aware training for large language models.
\newblock \emph{arXiv preprint arXiv:2305.17888}, 2023{\natexlab{c}}.

\bibitem[Lu et~al.(2023)Lu, Peng, Cheng, Galley, Chang, Wu, Zhu, and Gao]{lu2023chameleon}
Pan Lu, Baolin Peng, Hao Cheng, Michel Galley, Kai-Wei Chang, Ying~Nian Wu, Song-Chun Zhu, and Jianfeng Gao.
\newblock Chameleon: Plug-and-play compositional reasoning with large language models.
\newblock \emph{arXiv preprint arXiv:2304.09842}, 2023.

\bibitem[Men et~al.(2024)Men, Xu, Zhang, Wang, Lin, Lu, Han, and Chen]{men2024shortgpt}
Xin Men, Mingyu Xu, Qingyu Zhang, Bingning Wang, Hongyu Lin, Yaojie Lu, Xianpei Han, and Weipeng Chen.
\newblock Shortgpt: Layers in large language models are more redundant than you expect.
\newblock \emph{arXiv preprint arXiv:2403.03853}, 2024.

\bibitem[Nallapati et~al.(2016)Nallapati, Zhou, Gulcehre, Xiang, et~al.]{nallapati2016abstractive}
Ramesh Nallapati, Bowen Zhou, Caglar Gulcehre, Bing Xiang, et~al.
\newblock Abstractive text summarization using sequence-to-sequence rnns and beyond.
\newblock \emph{arXiv preprint arXiv:1602.06023}, 2016.

\bibitem[Qian et~al.(2023)Qian, Tang, Yang, Liang, and Liu]{qian2023can}
Chen Qian, Huayi Tang, Zhirui Yang, Hong Liang, and Yong Liu.
\newblock Can large language models empower molecular property prediction?
\newblock \emph{arXiv preprint arXiv:2307.07443}, 2023.

\bibitem[Qin et~al.(2023)Qin, Zhang, Zhang, Chen, Yasunaga, and Yang]{qin2023chatgpt}
Chengwei Qin, Aston Zhang, Zhuosheng Zhang, Jiaao Chen, Michihiro Yasunaga, and Diyi Yang.
\newblock Is chatgpt a general-purpose natural language processing task solver?
\newblock \emph{arXiv preprint arXiv:2302.06476}, 2023.

\bibitem[Rajpurkar et~al.(2016)Rajpurkar, Zhang, Lopyrev, and Liang]{rajpurkar2016squad}
Pranav Rajpurkar, Jian Zhang, Konstantin Lopyrev, and Percy Liang.
\newblock Squad: 100,000+ questions for machine comprehension of text.
\newblock \emph{arXiv preprint arXiv:1606.05250}, 2016.

\bibitem[Ram et~al.(2023)Ram, Levine, Dalmedigos, Muhlgay, Shashua, Leyton-Brown, and Shoham]{ram2023context}
Ori Ram, Yoav Levine, Itay Dalmedigos, Dor Muhlgay, Amnon Shashua, Kevin Leyton-Brown, and Yoav Shoham.
\newblock In-context retrieval-augmented language models.
\newblock \emph{arXiv preprint arXiv:2302.00083}, 2023.

\bibitem[Sawada et~al.(2023)Sawada, Paleka, Havrilla, Tadepalli, Vidas, Kranias, Nay, Gupta, and Komatsuzaki]{sawada2023arb}
Tomohiro Sawada, Daniel Paleka, Alexander Havrilla, Pranav Tadepalli, Paula Vidas, Alexander Kranias, John~J Nay, Kshitij Gupta, and Aran Komatsuzaki.
\newblock Arb: Advanced reasoning benchmark for large language models.
\newblock \emph{arXiv preprint arXiv:2307.13692}, 2023.

\bibitem[Schuster et~al.(2022)Schuster, Fisch, Gupta, Dehghani, Bahri, Tran, Tay, and Metzler]{calm}
Tal Schuster, Adam Fisch, Jai Gupta, Mostafa Dehghani, Dara Bahri, Vinh~Q. Tran, Yi~Tay, and Donald Metzler.
\newblock Confident adaptive language modeling, 2022.

\bibitem[Sheng et~al.(2023)Sheng, Zheng, Yuan, Li, Ryabinin, Fu, Xie, Chen, Barrett, Gonzalez, et~al.]{sheng2023high}
Ying Sheng, Lianmin Zheng, Binhang Yuan, Zhuohan Li, Max Ryabinin, Daniel~Y Fu, Zhiqiang Xie, Beidi Chen, Clark Barrett, Joseph~E Gonzalez, et~al.
\newblock High-throughput generative inference of large language models with a single gpu.
\newblock \emph{arXiv preprint arXiv:2303.06865}, 2023.

\bibitem[Sun et~al.(2022)Sun, Liu, Zhu, Geng, Wu, He, Ni, Xie, Huang, and Qiu]{hash-skip-sun}
Tianxiang Sun, Xiangyang Liu, Wei Zhu, Zhichao Geng, Lingling Wu, Yilong He, Yuan Ni, Guotong Xie, Xuanjing Huang, and Xipeng Qiu.
\newblock A simple hash-based early exiting approach for language understanding and generation, 2022.

\bibitem[Tang et~al.(2023)Tang, Zhu, Li, Appalaraju, Mahadevan, and Manmatha]{tang2023deed}
Peng Tang, Pengkai Zhu, Tian Li, Srikar Appalaraju, Vijay Mahadevan, and R.~Manmatha.
\newblock Deed: Dynamic early exit on decoder for accelerating encoder-decoder transformer models, 2023.

\bibitem[Touvron et~al.(2023)Touvron, Lavril, Izacard, Martinet, Lachaux, Lacroix, Rozi{\`e}re, Goyal, Hambro, Azhar, et~al.]{touvron2023llama}
Hugo Touvron, Thibaut Lavril, Gautier Izacard, Xavier Martinet, Marie-Anne Lachaux, Timoth{\'e}e Lacroix, Baptiste Rozi{\`e}re, Naman Goyal, Eric Hambro, Faisal Azhar, et~al.
\newblock Llama: Open and efficient foundation language models.
\newblock \emph{arXiv preprint arXiv:2302.13971}, 2023.

\bibitem[Wang et~al.(2023)Wang, Chen, Chen, Wu, Zhu, Zeng, Luo, Lu, Zhou, Qiao, et~al.]{wang2023visionllm}
Wenhai Wang, Zhe Chen, Xiaokang Chen, Jiannan Wu, Xizhou Zhu, Gang Zeng, Ping Luo, Tong Lu, Jie Zhou, Yu~Qiao, et~al.
\newblock Visionllm: Large language model is also an open-ended decoder for vision-centric tasks.
\newblock \emph{arXiv preprint arXiv:2305.11175}, 2023.

\bibitem[Xiao et~al.(2023)Xiao, Tian, Chen, Han, and Lewis]{xiao2023efficient}
Guangxuan Xiao, Yuandong Tian, Beidi Chen, Song Han, and Mike Lewis.
\newblock Efficient streaming language models with attention sinks.
\newblock \emph{arXiv preprint arXiv:2309.17453}, 2023.

\bibitem[Xin et~al.(2020)Xin, Tang, Lee, Yu, and Lin]{deebert}
Ji~Xin, Raphael Tang, Jaejun Lee, Yaoliang Yu, and Jimmy Lin.
\newblock {D}ee{BERT}: Dynamic early exiting for accelerating {BERT} inference.
\newblock In Dan Jurafsky, Joyce Chai, Natalie Schluter, and Joel Tetreault (eds.), \emph{Proceedings of the 58th Annual Meeting of the Association for Computational Linguistics}, pp.\  2246--2251, Online, July 2020. Association for Computational Linguistics.
\newblock \doi{10.18653/v1/2020.acl-main.204}.
\newblock URL \url{https://aclanthology.org/2020.acl-main.204}.

\bibitem[Ye et~al.(2023)Ye, Zhang, Wang, Xu, and Zhang]{ye2023natural}
Ruosong Ye, Caiqi Zhang, Runhui Wang, Shuyuan Xu, and Yongfeng Zhang.
\newblock Natural language is all a graph needs.
\newblock \emph{arXiv preprint arXiv:2308.07134}, 2023.

\bibitem[Yin et~al.(2023{\natexlab{a}})Yin, Jaiswal, Liu, Kundu, and Wang]{Yin2023PruningSP}
Lu~Yin, Ajay Jaiswal, Shiwei Liu, Souvik Kundu, and Zhangyang Wang.
\newblock Pruning small pre-trained weights irreversibly and monotonically impairs "difficult" downstream tasks in llms.
\newblock 2023{\natexlab{a}}.
\newblock URL \url{https://api.semanticscholar.org/CorpusID:263620664}.

\bibitem[Yin et~al.(2023{\natexlab{b}})Yin, Wu, Zhang, Hsieh, Wang, Jia, Pechenizkiy, Liang, Wang, and Liu]{yin2023outlier}
Lu~Yin, You Wu, Zhenyu Zhang, Cheng-Yu Hsieh, Yaqing Wang, Yiling Jia, Mykola Pechenizkiy, Yi~Liang, Zhangyang Wang, and Shiwei Liu.
\newblock Outlier weighed layerwise sparsity (owl): A missing secret sauce for pruning llms to high sparsity.
\newblock \emph{arXiv preprint arXiv:2310.05175}, 2023{\natexlab{b}}.

\bibitem[Zhang et~al.(2022)Zhang, Roller, Goyal, Artetxe, Chen, Chen, Dewan, Diab, Li, Lin, et~al.]{zhang2022opt}
Susan Zhang, Stephen Roller, Naman Goyal, Mikel Artetxe, Moya Chen, Shuohui Chen, Christopher Dewan, Mona Diab, Xian Li, Xi~Victoria Lin, et~al.
\newblock Opt: Open pre-trained transformer language models.
\newblock \emph{arXiv preprint arXiv:2205.01068}, 2022.

\bibitem[Zhangheng et~al.(2023)Zhangheng, Liu, Chen, JAISWAL, Zhang, Wang, Krishnamoorthi, Chang, and Wang]{zhangheng2023sparse}
LI~Zhangheng, Shiwei Liu, Tianlong Chen, AJAY~KUMAR JAISWAL, Zhenyu Zhang, Dilin Wang, Raghuraman Krishnamoorthi, Shiyu Chang, and Zhangyang Wang.
\newblock Sparse cocktail: Every sparse pattern every sparse ratio all at once.
\newblock 2023.

\bibitem[Zheng et~al.(2023)Zheng, Chiang, Sheng, Zhuang, Wu, Zhuang, Lin, Li, Li, Xing, et~al.]{zheng2023judging}
Lianmin Zheng, Wei-Lin Chiang, Ying Sheng, Siyuan Zhuang, Zhanghao Wu, Yonghao Zhuang, Zi~Lin, Zhuohan Li, Dacheng Li, Eric Xing, et~al.
\newblock Judging llm-as-a-judge with mt-bench and chatbot arena.
\newblock \emph{arXiv preprint arXiv:2306.05685}, 2023.

\bibitem[Zhu(2021)]{zhu-2021-leebert}
Wei Zhu.
\newblock {L}ee{BERT}: Learned early exit for {BERT} with cross-level optimization.
\newblock In Chengqing Zong, Fei Xia, Wenjie Li, and Roberto Navigli (eds.), \emph{Proceedings of the 59th Annual Meeting of the Association for Computational Linguistics and the 11th International Joint Conference on Natural Language Processing (Volume 1: Long Papers)}, pp.\  2968--2980, Online, August 2021. Association for Computational Linguistics.
\newblock \doi{10.18653/v1/2021.acl-long.231}.
\newblock URL \url{https://aclanthology.org/2021.acl-long.231}.

\bibitem[Zhuo(2023)]{zhuo2023large}
Terry~Yue Zhuo.
\newblock Large language models are state-of-the-art evaluators of code generation.
\newblock \emph{arXiv preprint arXiv:2304.14317}, 2023.

\end{thebibliography}
\bibliographystyle{colm2024_conference}


\end{document}